\documentclass[conference,compsoc]{IEEEtran}
\IEEEoverridecommandlockouts
\usepackage[T1]{fontenc}
\usepackage{microtype}
\usepackage{amssymb}
\usepackage{cancel}
\usepackage[sort]{cite}
\usepackage[hyphens]{url}
\usepackage{float}
\usepackage{booktabs}
\usepackage{xcolor}
\usepackage{colortbl}
\usepackage{array}
\newcolumntype{R}{>{\raggedleft\arraybackslash}p{0.6cm}}
\usepackage{setspace}
\usepackage{longtable}   % table spans multiple pages
\usepackage{graphicx}
\usepackage[dvipsnames]{xcolor}
\usepackage{balance}

\usepackage[hidelinks]{hyperref}
\usepackage[capitalize,nameinlink,nosort]{cleveref}

\usepackage{xspace}
\usepackage[per-mode=symbol,mode=text,detect-all]{siunitx}
\DeclareSIUnit{\million}{M\@}
\DeclareSIUnit{\thousand}{k\@}
\usepackage[inline]{enumitem}
\usepackage{booktabs}
\usepackage{tabularx}
\usepackage{multirow}

\newif\ifanonymous
\newif\ifauthor
\newif\ifdraft

\anonymoustrue
\anonymousfalse

\authortrue
\authorfalse

\drafttrue
\draftfalse

\ifdraft
\newcommand{\todo}[1]{{\color{red}{\textbf{TODO:} #1}}}

\newcommand{\jp}[1]{{\color{magenta}{#1}}}
\newcommand{\jl}[1]{{\color{blue}{#1}}}
\newcommand{\arj}[1]{{\color{olive}{#1}}}
\newcommand{\pet}[1]{{\color{violet}{#1}}}
\newcommand{\sop}[1]{{\color{brown}{#1}}}
\else
\newcommand{\todo}[1]{}

\newcommand{\jp}[1]{{}}
\newcommand{\jl}[1]{{}}
\newcommand{\arj}[1]{{}}
\newcommand{\pet}[1]{{}}
\newcommand{\sop}[1]{{}}
\fi

\newcounter{dilemma}

\newcommand{\dilemmaref}[1]{\underline{\ref{#1}}\xspace}

\usepackage{pifont}

\usepackage{tcolorbox}
\tcbuselibrary{skins,breakable}

  {\end{tcolorbox}}

\ifauthor
    \usepackage{tikz}

\fi

\def\BibTeX{{\rm B\kern-.05em{\sc i\kern-.025em b}\kern-.08em
	T\kern-.1667em\lower.7ex\hbox{E}\kern-.125emX}}

\begin{document}

% From the CfP: The page length limits include the space allowed for all tables and figures. References, LLM Usage Statement, and appendices are excluded from page limits; reviewers are not required to view the appendices when evaluating submissions

% Shorten lists of author names
% \bstctlcite{IEEEexample:BSTcontrol}
% Disable ugly dashes that replace the list of authors
\bstctlcite{IEEEexample:BSTcontrolLong}

%\title{\fix{Working Title:} \chk{How Ethical is Analyzing the arXiv?\\ {\vspace{-0.5em} \large Dissecting and Analyzing an ``Immutable'' Live Dataset}}}
% JL: ``Immutable'' -- Is that really true?
% JL: ``Live'' -- Other Internet data too?
\title{What Patients Say They Can’t Access: Mining Cancer Screening Barriers from Online Communities}

\ifanonymous
	\author{%
		Anonymous Author(s)\\
		$\left[\text{Experience Paper}\right]$
	}
\else
	\author{
	\IEEEauthorblockN{%
		Ruimin Sun\IEEEauthorrefmark{1},
		Aokun Chen\IEEEauthorrefmark{2},
		Arjun Arunasalam\IEEEauthorrefmark{1}%
	}
	\IEEEauthorblockA{\IEEEauthorrefmark{1}Florida International University, Miami, FL, US $\cdot$
	\{rsun, aarunasa\}@fiu.edu}
	\IEEEauthorblockA{\IEEEauthorrefmark{2} Florida State University, Tallahassee, FL, US $\cdot$
	ac25cv@fsu.edu}
	}
\fi
\maketitle

\begin{abstract}
Cancer screening disparities persist across the U.S., yet the  patient-reported barriers driving low uptake remain poorly characterized  at scale.
While clinical surveys capture structured responses, they miss  the organic, unsolicited discourse where patients candidly describe real-world obstacles. 
In this paper, we conduct a large-scale analysis of over $56{,}000$ posts across six cancer-specific Reddit communities to surface barriers to cancer screening among U.S. patients. 
We apply a  pipeline combining keyword filtering, barrier triage, open coding, and  codebook synthesis to systematically extract and categorize  patient-reported barriers from 576 U.S.-located posts. 
Our qualitative analysis reveals recurring structural, informational, and psychosocial barriers that vary meaningfully by cancer type and community context. 
These findings lay the groundwork for future work aimed at proactively  identifying and addressing barriers at scale—spanning a broader range  of cancer types and populations—and informing the design of targeted  interventions, navigator programs, and health system policies that  translate patient-reported obstacles into actionable solutions.
\end{abstract}
% \begin{IEEEkeywords}
% ethical considerations; remediability; sensitivity; privacy; awareness; research ethics; public datasets
% \end{IEEEkeywords}

% \todo{Contact chairs about double-blindness, we cannot really remove the reference to our previous paper, even though the list of authors will likely differ significantly.}
% Response: We think that your proposed approach is sound and it would be good to refer to your work [1] as anonymized and you can compare it with [2,3]. If possible you can try too remove the direct IDs. We do not think PC members will go beyond to purposefully try to find the authors.

%\todo{Do we want to cover/mention the legal situation as well?}
%\jl{Might be hard to generalize since these assessments likely differ globally.}

%\jp{I would avoid giving a definition of ethics and rather refer to prior work.}

\section{Introduction}
\label{sec:intro}

Cancer screening is one of the most effective tools available for reducing cancer mortality: early detection of breast, colorectal, lung, and prostate cancers substantially improves survival outcomes and reduces treatment burden~\cite{us2024screening,national2011reduced}. Yet uptake of recommended screening remains persistently uneven across the United States, with well-documented disparities by race, income, geography, and insurance status~\cite{aguiar2025barriers,messerman2026s,agunwamba2023barriers}. Understanding the specific barriers that prevent patients from accessing screening is essential for designing interventions that address root causes rather than surface-level correlates.

Existing research on screening barriers relies predominantly on structured
surveys and clinical interviews~\cite{torrez2024identifying,faguy2020challenges,salman2025patient}.  These methods capture responses to investigator-defined categories, are  resource-intensive to administer at scale, and are subject to social  desirability bias: respondents may underreport cost concerns, distrust of  providers, or prior negative experiences in the presence of a clinician or researcher. As a result, the lived texture of barrier experiences---the  specific insurance friction, the specialist two hours away, the provider who  dismissed a symptom as stress---is rarely captured in clinical  datasets~\cite{messerman2026s,aguiar2025barriers}. Social media discourse, by contrast, offers high ecological validity: patients describe barriers in their own words, in contexts of their own choosing, without the demand characteristics of a research encounter.

Social media platforms, and Reddit in particular, offer a complementary vantage point. Cancer-related subreddits host large, active communities where patients and caregivers discuss diagnoses, treatments, and access challenges in candid, self-initiated prose~\cite{currin2021understanding,garg2020information,tripathi2022examination}.  Because posts are unsolicited and pseudonymous, they are less subject to the  interviewer effects that suppress sensitive disclosures in formal research  settings~\cite{von2025medical,pierce2024identifying}. Prior work has established Reddit as a valid source of naturalistic health discourse for conditions ranging from mental health to chronic illness~\cite{de2014mental,schneider2025beyond}, but its potential for characterizing cancer \textit{screening} barriers specifically---as distinct from treatment or survivorship---remains largely unexplored. 

We ask: what barriers to cancer screening do U.S. patients report in online communities, and how can large-scale qualitative analysis of social media discourse surface and structure those barriers at scale?

In this paper, we present a systematic large-scale analysis of over 56,000 posts drawn from six cancer-specific subreddits to surface and categorize U.S. patient-reported barriers to cancer screening. We apply a multi-pass pipeline combining geographic filtering, two-tier keyword screening, and four-phase inductive qualitative coding~\cite{braun2006using,gioia2013seeking}  to distill a final analytical corpus of 217 barrier-containing posts. This  process yields a structured 43-code codebook spanning six barrier  domains---access and geography, provider attitudes, information and navigation,  financial and insurance, health system coordination, and patient-level  psychosocial factors---grounded in naturalistic patient language rather than investigator-imposed categories.

Our analysis reveals that access and geographic barriers---particularly scheduling delays and travel distance---are the most frequently reported obstacles, followed closely by provider attitude barriers and psychosocial factors such as fear and avoidance. Barriers rarely appear in isolation: 85.7\% of coded posts contained multiple co-occurring codes, underscoring the compounding nature of screening obstacles~\cite{messerman2026s,aguiar2025barriers}.  These findings point toward multi-domain intervention designs and establish  a patient-centered evidence base for future work.

This work makes three contributions. First, we leverage NLP and large-scale social media analysis to surface patient-reported cancer screening barriers that are rarely visible in clinical datasets, demonstrating the feasibility of unsolicited online discourse as a complementary evidence source. Second, we  produce a validated 43-code barrier codebook grounded in naturalistic patient language, suitable as ground truth for training downstream classifiers  targeting barrier detection in social media and clinical  text. Third, we outline a  research agenda for applying NLP at scale to identify at-risk patients,  support patient navigation programs, and  inform health system policy---translating patient-reported obstacles into actionable, computable signals.
\section{Background and Related Work}
\label{sec:background}

\subsection{Cancer Screening Barriers}
Cancer screening disparities are well documented across income, race, geography, and insurance status~\cite{aguiar2025barriers, messerman2026s, agunwamba2023barriers}.  Survey-based studies using instruments such as the Behavioral Risk Factor Surveillance System (BRFSS) and the National Health Interview Survey (NHIS)  have identified cost, access, and provider recommendation as primary barriers  to mammography, colonoscopy, and lung cancer screening~\cite{torrez2024identifying, faguy2020challenges, salman2025patient, leopold2023barriers, belachew2025facilitators}. However, structured instruments capture responses only to predefined categories,  are subject to social desirability effects, and cannot represent the compounding,  multi-domain barrier combinations that patients describe in open-ended  discourse~\cite{messerman2026s, aguiar2025barriers}. Patient navigation interventions have demonstrated efficacy in reducing these barriers~\cite{pilar2025characteristics}, yet their design continues to rely on aggregate survey data rather than granular patient voice.

\subsection{Social Media as a Health Discourse Resource}
Online health communities, and Reddit in particular, have emerged as rich sources of unsolicited, naturalistic patient discourse~\cite{currin2021understanding, garg2020information, tripathi2022examination}. Reddit's pseudonymity and asynchronous format lower social desirability pressure and encourage elaboration on sensitive topics including cost, distrust, and prior negative experiences~\cite{von2025medical, pierce2024identifying}. Prior NLP  research has leveraged Reddit for mental health surveillance~\cite{deiner2024large}  and chronic illness characterization~\cite{schneider2025beyond, fairie2021categorising},  establishing methodological groundwork for the current study. Cancer communities on Reddit are large and active, but prior work has focused predominantly on treatment experience and survivorship rather than screening  access~\cite{spiegel2025exploring, tripathi2022examination, currin2021understanding}.

\subsection{NLP for Health Barriers and Qualitative Grounding}
Computational methods for extracting social determinants of health (SDOH) from clinical notes and patient narratives have advanced substantially, with annotated corpora and transformer-based classifiers now targeting housing, food security, and transportation barriers in EHR text~\cite{lybarger20232022, romanowski2023extracting, fairie2021categorising}.  Barrier extraction from social media remains less developed, and no prior work has produced a validated codebook for cancer screening barriers specifically. Our approach is grounded in established inductive qualitative 
traditions~\cite{braun2006using, gioia2013seeking}: single-author coding proceeds to thematic saturation before any automated classifier is trained,  positioning the codebook itself as the primary deliverable and as ground truth  for future supervised learning. Recent work on LLM-assisted thematic analysis further informs our pipeline design, demonstrating that computationally supported qualitative coding can achieve reproducible, human-validated outputs at scale~\cite{deiner2024large, wang2025lata, jafarian2026human}.
\section{Methodology}
\label{sec:method}

\begin{figure*}[th!]
  \centering
  \includegraphics[width=\textwidth]{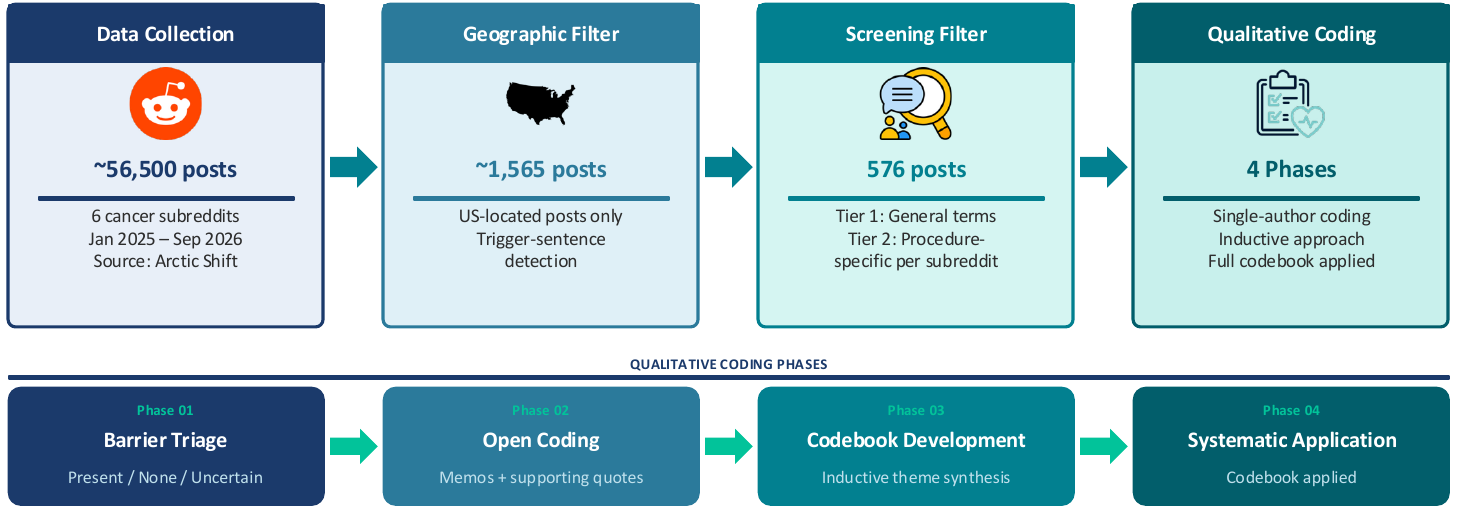}
  \caption{Study methodology overview. The data pipeline (top) shows the
           four filtering stages applied to \textasciitilde56{,}500 raw Reddit posts,
           yielding a final corpus of 576 screening-relevant posts.
           The qualitative coding phases (bottom) detail the four-phase
           inductive analysis conducted by a single author.}
  \label{fig:methodology}
\end{figure*}

Figure~\ref{fig:methodology} overviews our methodology as detailed below. 

\subsection{Data Collection}

We collected data from Reddit, a large-scale social media platform  organized into topic-specific communities (subreddits). Reddit users frequently discuss  personal medical experiences, concerns, and barriers in ways that  clinical surveys cannot easily elicit, making it well-suited for  exploratory qualitative inquiry into patient-reported experiences.

Subreddit selection was grounded in U.S. cancer incidence data. Drawing  on national cancer statistics \cite{cancerStats}, we identified  the leading cancer diagnoses by new case volume in the United States and  selected the corresponding subreddit for each: r/breastcancer, r/prostatecancer, r/lungcancer, r/colorectalcancer, and  r/pancreaticcancer. We additionally included r/Cancer as a  general-purpose community to capture discourse not tied to a single  diagnosis. Posts were collected spanning January 2025 through September  2026 using Arctic Shift~\cite{arcticshift2026}, a large-scale  Reddit archiving service. This yielded approximately 56,500 raw posts  across the six communities.

\subsection{Data Filtering}

\noindent \textbf{Geographic filtering.} Because this study targets barriers  experienced by U.S.-based patients, we applied a geographic filter to retain only posts containing explicit references to the United States~\cite{geonamescache}.  Specifically, we flagged posts containing U.S.-identifying trigger terms (e.g., state names, insurance terminology, U.S.-specific institutions)  within the post title, body, or top comments. This step reduced the  corpus to approximately 1,565 U.S.-located posts (2--4\% per subreddit).

\noindent \textbf{Screening relevance filtering.} We further filtered posts for  relevance to cancer screening using a two-tier keyword strategy. Tier 1  comprised general screening terms applicable across cancer types (e.g.,  \textit{``screening''}, \textit{``early detection''}, \textit{``routine test''}).  Tier 2 comprised procedure-specific terms tailored to each subreddit  (e.g., \textit{``mammogram''} for r/breastcancer, \textit{``colonoscopy''} for r/colorectalcancer, \textit{``PSA test''} for r/prostatecancer, \textit{``low-dose CT''} for r/lungcancer). A post was retained if it matched at least one term from either tier. This process yielded a final analytic corpus of  576 posts, with representation across all six communities. A full list of keywords is provided in Appendix A.1.

\subsection{Qualitative Coding}

Each post was coded as a composite unit comprising the post title, body and text, providing sufficient context to  assess whether a barrier was present and to extract supporting evidence.

\noindent \textbf{Phase 1 -- Barrier triage.} Each post was first assessed for whether it contained evidence of a screening barrier, resulting in a  three-way classification: \textit{barrier present}, \textit{no barrier},  or \textit{uncertain}. Posts classified as no barrier were excluded from  subsequent phases.

\noindent \textbf{Phase 2 -- Open coding.} For each barrier-containing post, open codes and analytic memos were generated, capturing the nature of the  barrier in the author's own words alongside a verbatim supporting quote.  This phase followed an inductive approach \cite{braun2006using},  allowing categories to emerge from the data rather than from a  predetermined framework.

\noindent \textbf{Phase 3 -- Codebook development.} Memos and open codes from  Phase 2 were aggregated and reviewed holistically to identify recurring  themes and consolidate overlapping codes. This synthesis produced a  structured codebook defining barrier categories, inclusion criteria, and  illustrative exemplars.

\noindent \textbf{Phase 4 -- Systematic coding.} The finalized codebook was  applied systematically across the full corpus of barrier-containing  posts.

\subsection{Rigor and Reflexivity}

All qualitative coding was conducted by a single author. Coding continued until thematic saturation was reached, with no new barrier categories emerging in the final 10\% of coded posts. The author's positionality as a computer scientist engaging in qualitative  health research is acknowledged as a potential source of interpretive  bias, and analytic decisions were guided by established standards for qualitative rigor.
\section{Findings}
\label{sec:finding}

\begin{figure}[t]
  \centering
  \includegraphics[width=\columnwidth]{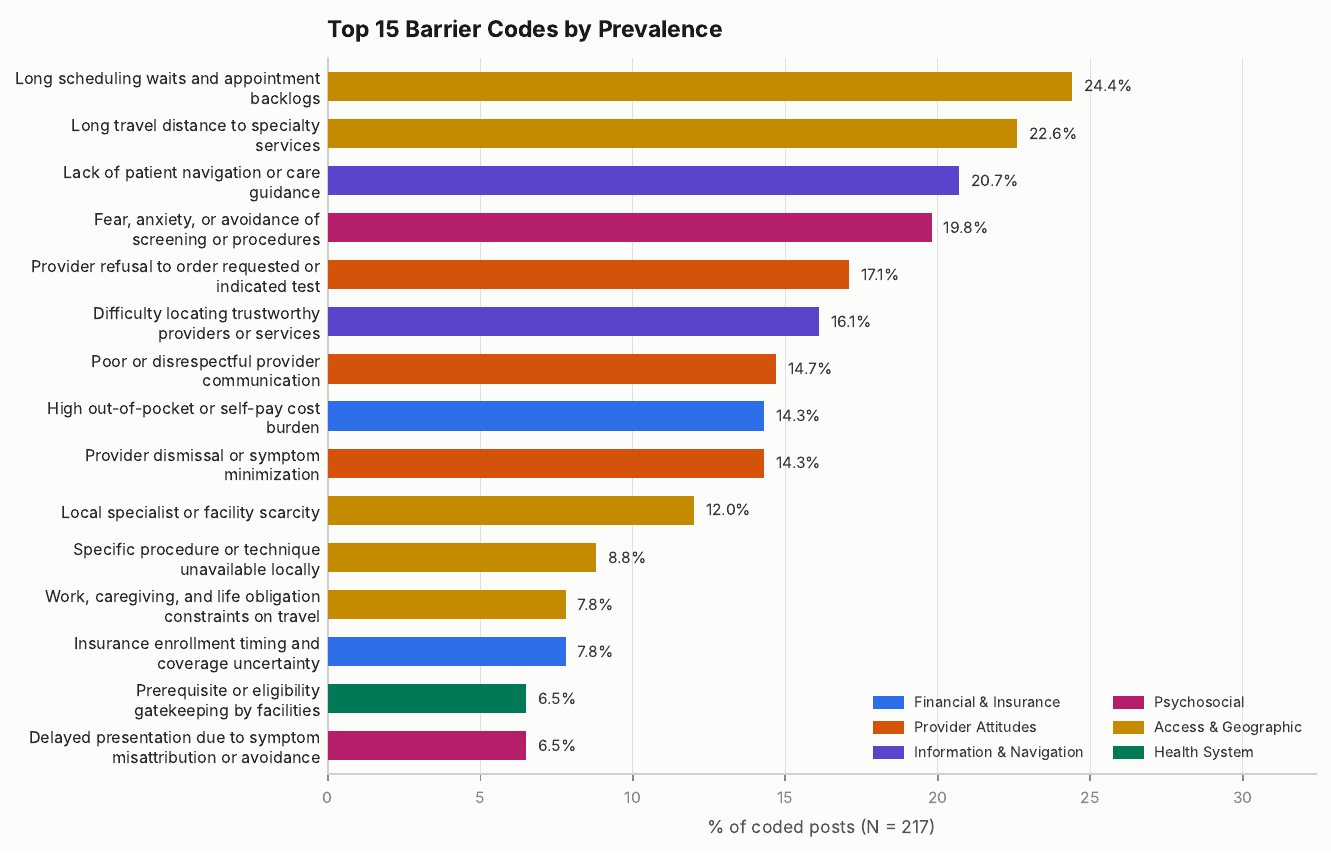}
  \caption{The 15 most prevalent cancer screening barrier codes identified
           through inductive qualitative coding of 217 Reddit posts. Bars are colored by domain:
           Access \& Geographic barriers (gold) account for five of the top
           15 codes, with scheduling delays (24.4\%) and travel distance
           (22.6\%) ranking first and second. Provider attitude barriers
           (orange) and Information \& Navigation barriers (indigo) each
           contribute multiple codes in the top ten, underscoring that
           barriers rarely operate in isolation.}
  \label{fig:top15}
\end{figure}

Of the 576 posts retained after two-tier keyword screening, 217 (37.7\%) were assigned at least one barrier code following Phase~4 systematic application of the finalized 43-code codebook. These 217 posts constitute the analytical corpus. Coding averaged 2.94 codes per post; 85.7\% of coded posts received two or more codes, reflecting the compounding, multi-domain nature of the barriers described. Approximately 10.6\% of posts were flagged as uncertain, and in 17.5\% of cases the coder noted downstream resolution, where the poster ultimately overcame the barrier through advocacy, travel, or repeated provider contact. Barriers were distributed across six domains: Access, Geographic, and Capacity Barriers (D05); Provider Attitudes, Decisions, and Communication (D02); Health System Processes, Coordination, and Policy (D06); Financial and Insurance Barriers (D01); Information and Navigation Barriers (D03); and Patient-Level Psychosocial and Emotional Barriers (D04). All illustrative excerpts below are close paraphrases of source posts, verified to preserve the semantic content and analytical meaning of the original text while mitigating re-identification risk~\cite{ayers2018don}. Figure~\ref{fig:top15} presents the fifteen most prevalent codes by percentage of coded posts (Detailed Breakdown in Appendix A.2).

% ──────────────────────────────────────────────────────────────────────────────
\subsection{Access, Geographic, and Capacity Barriers (D05)}

Access and geographic barriers were the dominant domain, accounting for 27.6\% of posts by primary code and appearing in five of the top fifteen codes. The most prevalent barrier in the entire dataset was \textit{long scheduling waits and appointment backlogs} (D05-C04), present in 24.4\% of coded posts. Posters described waits of weeks to months for biopsies, colonoscopies, and imaging, often exacerbated by cascading cancellations. One user recounted: \textit{``The snow delays in my area completely derailed my appointments---my biopsy and imaging got pushed back almost two months.''} Such delays were not isolated disruptions but systemic, reflecting facility-level capacity shortfalls rather than individual scheduling misfortunes.

\textit{Long travel distance to specialty services} (D05-C01) was the second most prevalent barrier overall (22.6\%), with posters routinely reporting drives of two or more hours to access procedures unavailable in their region. One colorectal cancer patient wrote: \textit{``The nearest facility that offers this procedure is the only one in the state, and it is well over two hours from where I live.''} This geographic scarcity was compounded by \textit{local specialist or facility scarcity} (D05-C03, 12.0\%) and \textit{specific procedure or technique unavailable locally} (D05-C06, 8.8\%), together painting a picture of concentrated specialty capacity in urban academic centers and thin coverage elsewhere. \textit{Work, caregiving, and life obligation constraints on travel} (D05-C05, 7.8\%) emerged as a structurally distinct access barrier: not that services were geographically unreachable, but that the burden of reaching them---multiple days off work, childcare, and lodging---made access functionally impossible for employed or caregiving patients.

% ──────────────────────────────────────────────────────────────────────────────
\subsection{Provider Attitudes, Decisions, and Communication (D02)}

Provider-related barriers constituted the second-largest domain by primary code (23.0\%). These posts captured a recurring pattern of clinical gatekeeping in which provider judgment, rather than evidence-based guidelines, determined whether screening proceeded. \textit{Provider refusal to order a requested or indicated test} (D02-C02, 17.1\%) was the fifth most prevalent code overall and encompassed cases where patients with significant symptom burden were denied workup. In one illustrative account, a provider attributed a patient's gastrointestinal symptoms to stress rather than ordering further workup: \textit{``He told me the bleeding and pain were probably just from stress and sent me home without any further testing.''} The same post described months of delayed diagnosis following repeated dismissals, a trajectory also captured by \textit{provider dismissal or symptom minimization} (D02-C01, 14.3\%).

A qualitatively distinct form of gatekeeping appeared in \textit{poor or disrespectful provider communication} (D02-C03, 14.7\%), where posters described scan denials framed as institutional protocol rather than clinical judgment: \textit{``They told me routine scans are not standard practice anymore---they only order them if a patient comes in with a specific complaint.''} This framing shifted interpretive responsibility away from the provider and onto the patient, leaving users uncertain whether their reluctance to push back was appropriate deference or missed advocacy. \textit{Provider practice constraints or convenience-driven limitations} (D02-C04, 6.5\%) captured cases where office policies---such as panel closures or preferred facility contracts---further constrained access independent of clinical need.

% ──────────────────────────────────────────────────────────────────────────────
\subsection{Information and Navigation Barriers (D03)}

Information and navigation barriers appeared in 9.2\% of posts by primary code but were among the most emotionally salient in the corpus. \textit{Lack of patient navigation or care guidance} (D03-C01) was the third most prevalent barrier overall (20.7\%), with posters describing confusion about how to initiate screening, where to go for a second opinion, or what their pathology results meant. A related barrier, \textit{difficulty locating trustworthy providers or services} (D03-C03, 16.1\%), reflected the effort patients expended vetting specialists and institutions, particularly after a negative experience with a prior provider.

A distinctive navigational barrier was \textit{failure to communicate or provide access to results} (D03-C02, 5.1\%), with institutional policies creating information bottlenecks between the patient and their own records. One poster described: \textit{``The office will not release my results to
anyone else until my oncologist reviews and signs off, which can take up to two weeks.''} Such delays prevented patients from seeking timely second
opinions and illustrate how information flow itself becomes a structural access barrier.

% ──────────────────────────────────────────────────────────────────────────────
\subsection{Financial and Insurance Barriers (D01)}

Financial and insurance barriers accounted for 15.7\% of posts by primary code. \textit{High out-of-pocket or self-pay cost burden} (D01-C01, 14.3\%) was the most direct expression of financial constraint, with posters describing stark variation in cost depending on facility choice: \textit{``The out-of-pocket at one facility was nearly double what another place would charge for the exact same procedure.''} For some, cost was prohibitive without any coverage: \textit{``I had no insurance and no money, so I walked into the ER and just asked for help.''}

\textit{Insurance enrollment timing and coverage uncertainty} (D01-C05, 7.8\%) captured a distinct administrative dimension, where gaps in coverage windows left patients unable to schedule procedures despite having a diagnosis. \textit{Prior authorization or insurer approval delay} (D01-C03, 6.0\%) and \textit{network or plan acceptance and contracting limits} (D01-C04, 6.0\%) together described the bureaucratic friction between the treating team's recommendation and payer approval, a friction that introduced weeks of latency into time-sensitive diagnostic pathways.

% ──────────────────────────────────────────────────────────────────────────────
\subsection{Health System Processes, Coordination, and Policy (D06)}

Health system barriers (16.1\% by primary code) captured structural failures in institutional coordination that were external to any single provider. \textit{Prerequisite or eligibility gatekeeping by facilities} (D06-C06, 6.5\%) described situations where institutions imposed procedural requirements before agreeing to perform screening, including requirements to establish care with a primary physician or to complete prior studies at the same institution. One poster described a circular conflict between payer and facility requirements: \textit{``They wanted me to redo the MRI through their system before they would move forward, even though I had just had one done elsewhere.''} \textit{Fragmented care across institutions requiring patient coordination} (D06-C07, 6.0\%) captured the downstream burden of this fragmentation, with patients themselves managing records transfers, repeat tests, and cross-site scheduling in the absence of institutional coordination. Additional codes---including \textit{pathology turnaround delay or conflicting interpretation} (D06-C08, 4.6\%) and \textit{failure to act on or communicate abnormal results} (D06-C05, 4.1\%)---reflect downstream delays that may occur after the procedure itself, extending the diagnostic gap into the post-test phase. One poster described: \textit{``My appointment was cancelled because apparently I needed to meet a requirement nobody told me about---now I am back to square one.''}

% ──────────────────────────────────────────────────────────────────────────────
\subsection{Patient-Level Psychosocial and Emotional Barriers (D04)}

Psychosocial barriers, while representing 7.8\% of posts by primary code, were  clinically significant in character. \textit{Fear, anxiety, or avoidance of screening or procedures} (D04-C01) was the fourth most prevalent code in the entire dataset (19.8\%), appearing nearly as frequently as structural barriers, and co-occurring with them in many posts. Posts in this code ranged from procedural anxiety---particularly around sedation and biopsy---to cancer dread. One poster described the compounding of prior medical trauma on a family member's willingness to undergo biopsy: \textit{``She has a lot of trauma around medical procedures, so even the idea of a biopsy brought up a lot of fear and made the whole process much harder.''} Another expressed overwhelming emotional burden accumulating across concurrent diagnostic workups: \textit{``Dealing with the possibility of multiple diagnoses at the same time has been completely overwhelming.''}

\textit{Delayed presentation due to symptom misattribution or avoidance} (D04-C02, 6.5\%) captured cases where patients waited weeks to months before seeking care, often normalizing symptoms as benign. These posts were notable because delay was self-initiated rather than system-imposed, though the co-occurrence pattern suggests that structural barriers---cost, access, provider dismissal---may reinforce initial avoidance by removing pathways that would have encouraged earlier engagement.

% ──────────────────────────────────────────────────────────────────────────────
\subsection{Cross-Cutting Patterns}

Two cross-cutting patterns emerged consistently across domains. First, barriers rarely appeared in isolation: 85.7\% of coded posts received two or more codes, and the combination of a geographic barrier with a financial barrier was among the most common pairings, reflecting the compounding cost of traveling to distant specialty centers on top of already high out-of-pocket burdens. Second, the corpus contained evidence of \textit{barrier-driven advocacy}, with 17.5\% of posts documenting that the poster had navigated around or ultimately resolved the barrier through persistent effort---seeking second opinions, traveling to distant centers, or leveraging social networks to locate providers. These resolution trajectories underscore that the barriers identified are not uniformly insurmountable but do impose a substantial burden of self-navigation onto patients, a burden distributed unevenly by geography, insurance status, and social capital.
\section{Discussion}
\label{sec:discussion}

The filtering pipeline's dramatic reduction from 56{,}000 raw posts to 576 screening-relevant posts reflects the inherently conversational nature of cancer subreddits: the majority of discourse centers on treatment experiences, emotional support, and prognosis rather than screening. That 576 posts nonetheless yielded a rich, codeable corpus suggests that when patients do discuss screening, they do so with specificity and depth---articulating systemic, financial, and psychological barriers in language that is difficult to elicit through structured instruments.

\subsection{Reddit as a Methodological Choice} Reddit offers several properties that distinguish it from conventional data sources. Posts are unsolicited and self-initiated, eliminating interviewer effects that can suppress sensitive disclosures around cost, distrust of the medical system, or prior negative experiences. The asynchronous format encourages elaboration, and the pseudonymous environment lowers social desirability pressure. Critically, ArcticShift provides access to longitudinal archives, enabling future researchers to track how screening barrier discourse evolves over time in response to policy changes or public health campaigns. The tradeoff is a population skewed toward English-speaking, internet-active individuals. This limitation may bound generalizability to the broader cancer patient population.

\subsection{Situating Qualitative Coding in an NLP Context}
Single-author inductive coding was deliberately chosen over automated classification at this stage. 
Building a reliable ground-truth codebook requires close, iterative engagement with raw text before any classifier can be meaningfully trained or evaluated. 
The four-phase process involving barrier triage, open coding, codebook development, and systematic application mirrors established qualitative traditions while producing structured outputs (coded labels, memos, uncertainty flags) that are directly amenable to downstream supervised learning. 
This positions the current work as a necessary precursor to scalable NLP rather than an alternative to it.

\subsection{Limitations}
Geographic filtering via trigger-sentence detection is a heuristic and will misclassify some non-US posts while excluding US posters who omit location cues.
Single-coder analysis, while appropriate for codebook development, precludes inter-rater reliability assessment; future iterations should incorporate a second coder on a random subset. Finally, Reddit's demographic skews may ignore barriers faced by older adults, rural populations, and non-English speakers that are likely underrepresented in this corpus.

\subsection{Broader Implications and Research Agenda}
Despite these limitations, the methodology demonstrates that structured qualitative analysis of social media archives can yield actionable, patient-centered insight at minimal cost relative to primary data collection. The resulting 43-code codebook has direct utility for public health researchers designing targeted outreach, hospital systems auditing screening program accessibility, and NLP practitioners developing barrier detection models for clinical decision support.

More broadly, this work establishes a foundation for a research agenda applying NLP at scale to cancer screening barriers. 
We identify three priority directions. 
First, the codebook provides labeled ground truth for training supervised classifiers capable of detecting and categorizing barriers in social media at scale, which would enable systematic monitoring of barrier discourse across Reddit archives and other online health communities without manual coding.
Second, the barrier taxonomy developed here is directly transferable to clinical text: transformer-based models trained on the codebook could be applied to EHR notes, patient portal messages, and navigator call logs to surface screening obstacles documented in clinical workflows but rarely aggregated or acted upon. 
Third, the co-occurrence structure of the codebook which reflect how barriers compound across domains opens pathways for multi-label barrier profiling of individual patients, which could inform the design of targeted navigation interventions that address the specific barrier combination a patient faces rather than treating barriers as independent. Together, these directions position the codebook not merely as a research artifact but as operational infrastructure for NLP systems aimed at reducing screening disparities at population scale.

\section{Conclusion}

This work demonstrates that Reddit constitutes a tractable corpus for surfacing cancer screening barriers that rarely appear in clinical documentation. Applying a two-tier filtering pipeline to over 56{,}000 posts across six cancer-specific subreddits, we distilled 576 posts into a structured codebook spanning access, awareness, fear, and system-level friction---grounded in naturalistic patient language rather than survey instruments.

The findings position social media NLP as a low-cost complement to traditional health services research, capturing unsolicited patient voice that formal settings often miss. Future work will extend this framework to longitudinal trend analysis and automated barrier classification scalable to broader Reddit archives.

% \ifanonymous
% \else
% \section*{Acknowledgment}
% \label{sec:acks}

% We appreciate the openness of and the valuable exchange with the computer security community when putting together this work:
% Fruitful discussions enabled this exhaustive work.
% %
% % Do not show this ack.
% We are further grateful for the valuable feedback by Sophie Stephenson\chk{ and Tadayoshi Kohno}.
% We also thank the anonymous reviewers for their valuable feedback and suggestions.

%\todo{This content must end on Page~8. Currently on \thepage.}

% % From the CfP: Authors are strongly encouraged to share artifacts whenever possible [...] This statement should be present during both submission and in the final version of the paper.
% \section*{Research Artifact Statement}
% This work did not produce any artifacts.
% Its contributions are of an intellectual nature; it discusses works that released their artifacts.
% Prior work~\cite{Pennekampetal2026Hidden,Dubniczkyetal2026You} released partial artifacts.

\balance
\bibliographystyle{IEEEtran}
\bibliography{paper}

\appendices
% appendix/keywords.tex
% Screening keyword tables — included from appendix.tex

\section*{A.1 Screening Keyword Lists}
\addcontentsline{toc}{section}{A.1\quad Screening Keyword Lists}

Posts were retained for downstream qualitative coding if they matched at
least one term from either Tier~1/Tier~2. 

% ── Tier 1 ───────────────────────────────────────────────────────────────────

\begin{table}[H]
\centering
\caption{Tier~1 keywords applied across all six subreddits.}
\label{tab:tier1-keywords}
\small
\begin{tabular}{ll}
\toprule
\textbf{Keyword} & \textbf{Keyword} \\
\midrule
screening            & early screening        \\
early detection      & cancer test            \\
preventive           & cancer check           \\
prevention           & cancer screening       \\
routine checkup      & follow-up test         \\
routine check-up     & detect early           \\
routine check        & annual exam            \\
annual check         &                        \\
\bottomrule
\end{tabular}
\end{table}

% ── Tier 2 ───────────────────────────────────────────────────────────────────

{\footnotesize\begin{table}[H]
\centering
\caption{Tier~2 keywords organized by cancer-type category.}
\label{tab:tier2-keywords}
\small
\begin{tabular}{ll}
\toprule
\textbf{Category} & \textbf{Keywords} \\
\midrule
\multirow{2}{*}{Breast}
  & mammogram, mammography, \\
  & brca, genetic test, genetic testing \\
\addlinespace
\multirow{3}{*}{Colorectal}
  & colonoscopy, stool test, fit test, fobt, \\
  & sigmoidoscopy, cologuard, \\
  & fecal immunochemical, fecal occult \\
\addlinespace
\multirow{2}{*}{Lung}
  & low-dose ct, low dose ct, ldct, \\
  & lung ct, lung scan \\
\addlinespace
\multirow{3}{*}{Prostate}
  & psa test, psa level, psa screening, \\
  & psa score, prostate exam, \\
  & digital rectal, dre \\
\addlinespace
\multirow{2}{*}{Pancreatic}
  & ca 19-9, ca19-9, \\
  & endoscopic ultrasound \\
\addlinespace
\multirow{3}{*}{General Procedures}
  & biopsy, ultrasound screening, \\
  & mri screening, pet scan, \\
  & ct screening, endoscopy \\
\bottomrule
\end{tabular}
\end{table}
}

% appendix/codebook_prevalence.tex
% Required packages: booktabs, xcolor, colortbl, longtable, array, multicol

\section*{A.2\quad Full Codebook Prevalence}
\addcontentsline{toc}{section}{A.2\quad Full Codebook Prevalence}

Table~\ref{tab:codebook-full} reports the frequency and percentage of coded
posts ($N = 217$) for all 43 barrier codes, grouped by domain. Shaded rows
mark domain boundaries; codes within each domain are ordered by prevalence.
Because posts receive multiple codes, percentages sum to more than 100\%.

\definecolor{domAccess}{HTML}{D4A843}
\definecolor{domProvider}{HTML}{E07B39}
\definecolor{domInfo}{HTML}{5B6DB8}
\definecolor{domFinancial}{HTML}{C0392B}
\definecolor{domSystem}{HTML}{27AE60}
\definecolor{domPsycho}{HTML}{8E44AD}
\definecolor{rowshade}{HTML}{F5F5F5}

\setlength{\extrarowheight}{2pt}

% longtable requires 1-column mode — temporarily switch if in twocolumn layout
\onecolumn

\begin{longtable}{@{}p{1.6cm}p{11.5cm}rr@{}}

\caption{All 43 barrier codes by domain and prevalence ($N=217$ coded posts).
Percentages exceed 100\% because posts receive multiple codes.}
\label{tab:codebook-full}\\

\toprule
\textbf{Code} & \textbf{Barrier Description} & \textbf{\textit{n}} & \textbf{\%} \\
\midrule
\endfirsthead

\multicolumn{4}{c}{\small\itshape Table~\ref{tab:codebook-full} continued}\\[4pt]
\toprule
\textbf{Code} & \textbf{Barrier Description} & \textbf{\textit{n}} & \textbf{\%} \\
\midrule
\endhead

\midrule
\multicolumn{4}{r}{\small\itshape Continued on next page}\\
\endfoot

\bottomrule
\endlastfoot

% ── D05: Access, Geographic, and Capacity Barriers ──────────────────────────
\rowcolor{domAccess!30}
\multicolumn{4}{l}{\quad\textbf{D05\enspace Access, Geographic, and Capacity Barriers}} \\
\midrule
D05-C04 & Long scheduling waits and appointment backlogs                         & 53 & 24.4\% \\
\rowcolor{rowshade}
D05-C01 & Long travel distance to specialty services                             & 49 & 22.6\% \\
D05-C03 & Local specialist or facility scarcity                                  & 26 & 12.0\% \\
\rowcolor{rowshade}
D05-C06 & Specific procedure or technique unavailable locally                    & 19 &  8.8\% \\
D05-C05 & Work, caregiving, and life obligation constraints on travel            & 17 &  7.8\% \\
\rowcolor{rowshade}
D05-C02 & Lack of personal transportation or reliance on others                  &  6 &  2.8\% \\
D05-C07 & Procedural sedation or pain management unavailability                  &  4 &  1.8\% \\
\rowcolor{rowshade}
D05-C08 & Clinical trial access barriers                                         &  3 &  1.4\% \\
D05-C09 & Referral catch-22 requiring unavailable prerequisite procedure         &  1 &  0.5\% \\
\addlinespace[6pt]

% ── D02: Provider Attitudes, Decisions, and Communication ───────────────────
\rowcolor{domProvider!30}
\multicolumn{4}{l}{\quad\textbf{D02\enspace Provider Attitudes, Decisions, and Communication}} \\
\midrule
D02-C02 & Provider refusal to order requested or indicated test                  & 37 & 17.1\% \\
\rowcolor{rowshade}
D02-C03 & Poor or disrespectful provider communication                           & 32 & 14.7\% \\
D02-C01 & Provider dismissal or symptom minimization                             & 31 & 14.3\% \\
\rowcolor{rowshade}
D02-C04 & Provider practice constraints or convenience-driven limitations        & 14 &  6.5\% \\
D02-C05 & Care continuity disruption from provider turnover                      &  5 &  2.3\% \\
\addlinespace[6pt]

% ── D03: Information and Navigation Barriers ────────────────────────────────
\rowcolor{domInfo!30}
\multicolumn{4}{l}{\quad\textbf{D03\enspace Information and Navigation Barriers}} \\
\midrule
D03-C01 & Lack of patient navigation or care guidance                            & 45 & 20.7\% \\
\rowcolor{rowshade}
D03-C03 & Difficulty locating trustworthy providers or services                  & 35 & 16.1\% \\
D03-C02 & Failure to communicate or provide access to results                    & 11 &  5.1\% \\
\rowcolor{rowshade}
D03-C04 & Cross-border or cross-jurisdiction information and record acceptance   &  5 &  2.3\% \\
\addlinespace[6pt]

% ── D01: Financial and Insurance Barriers ───────────────────────────────────
\rowcolor{domFinancial!30}
\multicolumn{4}{l}{\quad\textbf{D01\enspace Financial and Insurance Barriers}} \\
\midrule
D01-C01 & High out-of-pocket or self-pay cost burden                             & 31 & 14.3\% \\
\rowcolor{rowshade}
D01-C05 & Insurance enrollment timing and coverage uncertainty                   & 17 &  7.8\% \\
D01-C03 & Prior authorization or insurer approval delay                          & 13 &  6.0\% \\
\rowcolor{rowshade}
D01-C04 & Network or plan acceptance and contracting limits                      & 13 &  6.0\% \\
D01-C02 & Insurance coverage denial or non-covered classification                & 12 &  5.5\% \\
\rowcolor{rowshade}
D01-C06 & Policy-based frequency limits on surveillance imaging                  &  2 &  0.9\% \\
\addlinespace[6pt]

% ── D06: Health System Processes, Coordination, and Policy ──────────────────
\rowcolor{domSystem!30}
\multicolumn{4}{l}{\quad\textbf{D06\enspace Health System Processes, Coordination, and Policy}} \\
\midrule
D06-C06 & Prerequisite or eligibility gatekeeping by facilities                  & 14 &  6.5\% \\
\rowcolor{rowshade}
D06-C07 & Fragmented care across institutions requiring patient coordination     & 13 &  6.0\% \\
D06-C15 & Policy or regulatory constraints on coverage and access                & 11 &  5.1\% \\
\rowcolor{rowshade}
D06-C08 & Pathology turnaround delay or conflicting interpretation               & 10 &  4.6\% \\
D06-C05 & Failure to act on or communicate abnormal results                      &  9 &  4.1\% \\
\rowcolor{rowshade}
D06-C01 & Referral not sent or incorrectly processed                             &  6 &  2.8\% \\
D06-C09 & False negative or missed detection by standard screening               &  6 &  2.8\% \\
\rowcolor{rowshade}
D06-C14 & Public system or safety-net clinic overload                            &  5 &  2.3\% \\
D06-C11 & Duplicate testing or in-network re-verification requirement            &  4 &  1.8\% \\
\rowcolor{rowshade}
D06-C10 & Inconclusive biopsy or repeat biopsy requirement                       &  4 &  1.8\% \\
D06-C04 & Scheduling and booking errors                                          &  3 &  1.4\% \\
\rowcolor{rowshade}
D06-C02 & Medical records transfer delays or failures                            &  2 &  0.9\% \\
D06-C13 & Reliance on individual advocates to secure timely care                 &  2 &  0.9\% \\
\rowcolor{rowshade}
D06-C12 & Specimen transport or laboratory processing failure                    &  1 &  0.5\% \\
D06-C03 & Specimen, tissue block, or biopsy transfer barriers                    &  1 &  0.5\% \\
\addlinespace[6pt]

% ── D04: Patient-Level Psychosocial and Emotional Barriers ──────────────────
\rowcolor{domPsycho!30}
\multicolumn{4}{l}{\quad\textbf{D04\enspace Patient-Level Psychosocial and Emotional Barriers}} \\
\midrule
D04-C01 & Fear, anxiety, or avoidance of screening or procedures                 & 43 & 19.8\% \\
\rowcolor{rowshade}
D04-C02 & Delayed presentation due to symptom misattribution or avoidance       & 14 &  6.5\% \\
D04-C03 & Language barriers and limited health literacy                          &  4 &  1.8\% \\
\rowcolor{rowshade}
D04-C04 & Shame, stigma, or reluctance to seek care                              &  2 &  0.9\% \\

\end{longtable}

% Return to two-column layout after the longtable
\twocolumn

\end{document}